\documentclass{article}

\usepackage[nonatbib,final]{neurips_2019}
\usepackage[utf8]{inputenc} % allow utf-8 input
\usepackage[T1]{fontenc}    % use 8-bit T1 fonts
\usepackage{hyperref}       % hyperlinks
\usepackage{url}            % simple URL typesetting
\usepackage{booktabs}       % professional-quality tables
\usepackage{amsfonts}       % blackboard math symbols
\usepackage{nicefrac}       % compact symbols for 1/2, etc.
\usepackage{microtype}      % microtypography
\usepackage{graphicx}
\usepackage{amsmath}

\title{Application of Disentanglement to Map Registration Problem} 

\author{%
 Hae Jin Song~\thanks{Corresponding author: haejinso@usc.edu} \\
 Department of Computer Science\\
  University of Southern California\\
   \And
   Patrycja Krawczuk \\
   Department of Computer Science\\
   University of Southern California\\
    \AND
    Po-Hsuan Huang \\
    Neuroscience Graduate Program\\
    University of Southern California \\
}

\begin{document}

\maketitle

\begin{abstract}
Geospatial data come from various sources, such as satellites, aircraft, and LiDAR.  The variability of the source is not limited to the types of data acquisition techniques, as we have maps from different time periods. To incorporate these data for a coherent analysis, it is essential to first align different "styles" of geospatial data to its matching images that point to the same location on the surface of the Earth. In this paper, we approach the image registration as a two-step process of (1) extracting geospatial contents invariant to visual (and any other non-content-related) information, and (2) matching the data based on such (purely) geospatial contents. We hypothesize that a combination of $\beta$-VAE-like architecture~\cite{HigginsMPBGBML17} and adversarial training will achieve both  the disentanglement of the geographic information and artistic styles and  generation of new map tiles by composing the encoded geographic information with any artistic style.
\end{abstract}

\section{Motivation and Hypothesis}

\subsection{Motivation}
In Geographic Information Systems (GIS) and Remote Sensing, image registration refers to aligning images from different sources. It has a wide range of applications from image fusion, map digitization, geo-referencing, and globe-scale change detection to climate monitoring.  For example, we could have images from multi-band satellites and would like to align them on top of maps currently available on Google Maps or OpenStreetMap. Our project is motivated by another interesting application, which is to register historical maps with contemporary maps. By aligning maps from different periods, with vastly different style domains, we can gain a richer story about a region's history, as well as accurate information to guide future decision making.  
\begin{figure}[ht]
\centering
\begin{minipage}{.30\linewidth}
  \includegraphics[width = 4cm ,height = 4cm]{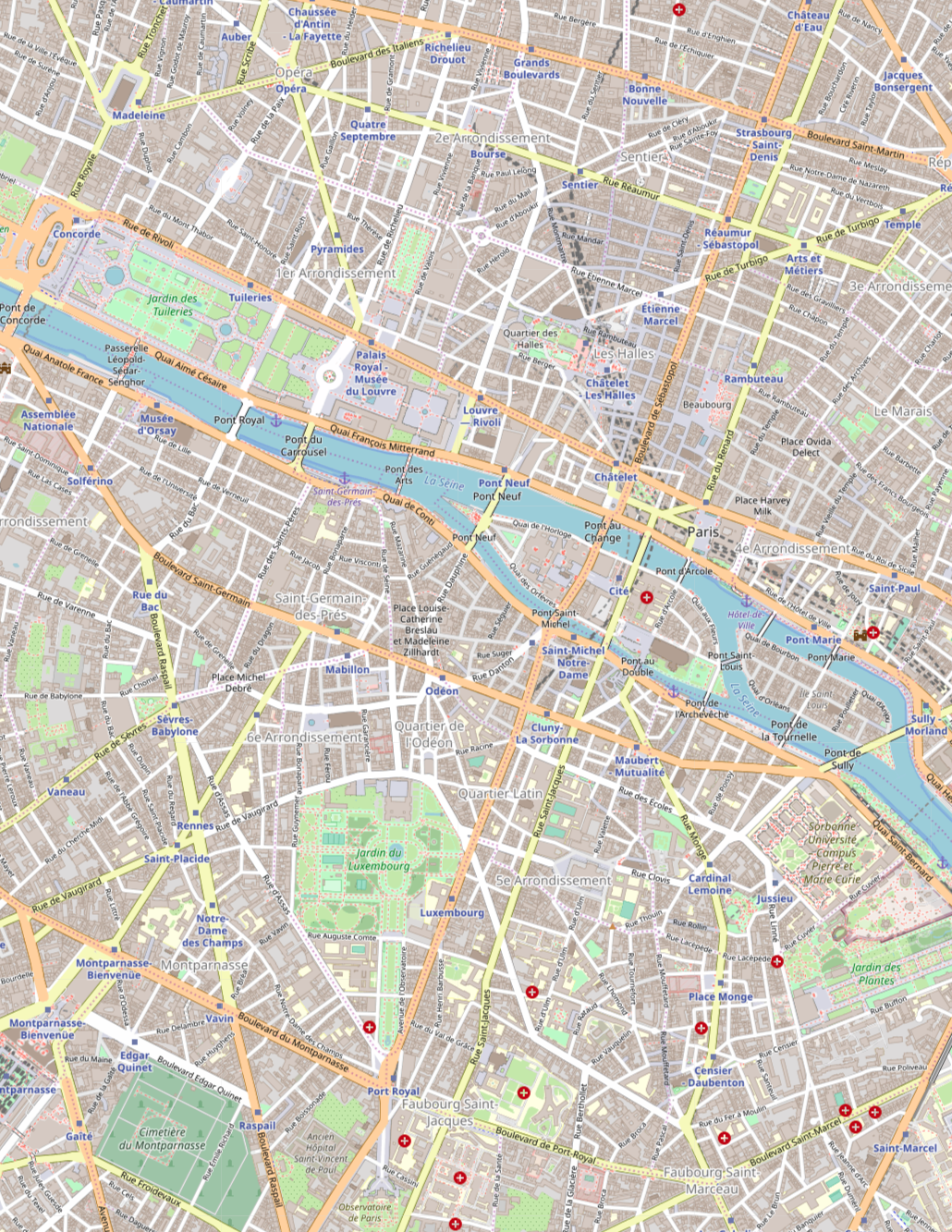}
\end{minipage}
\hspace{.05\linewidth}
\begin{minipage}{.30\linewidth}
  \includegraphics[width = 4cm ,height = 4cm]{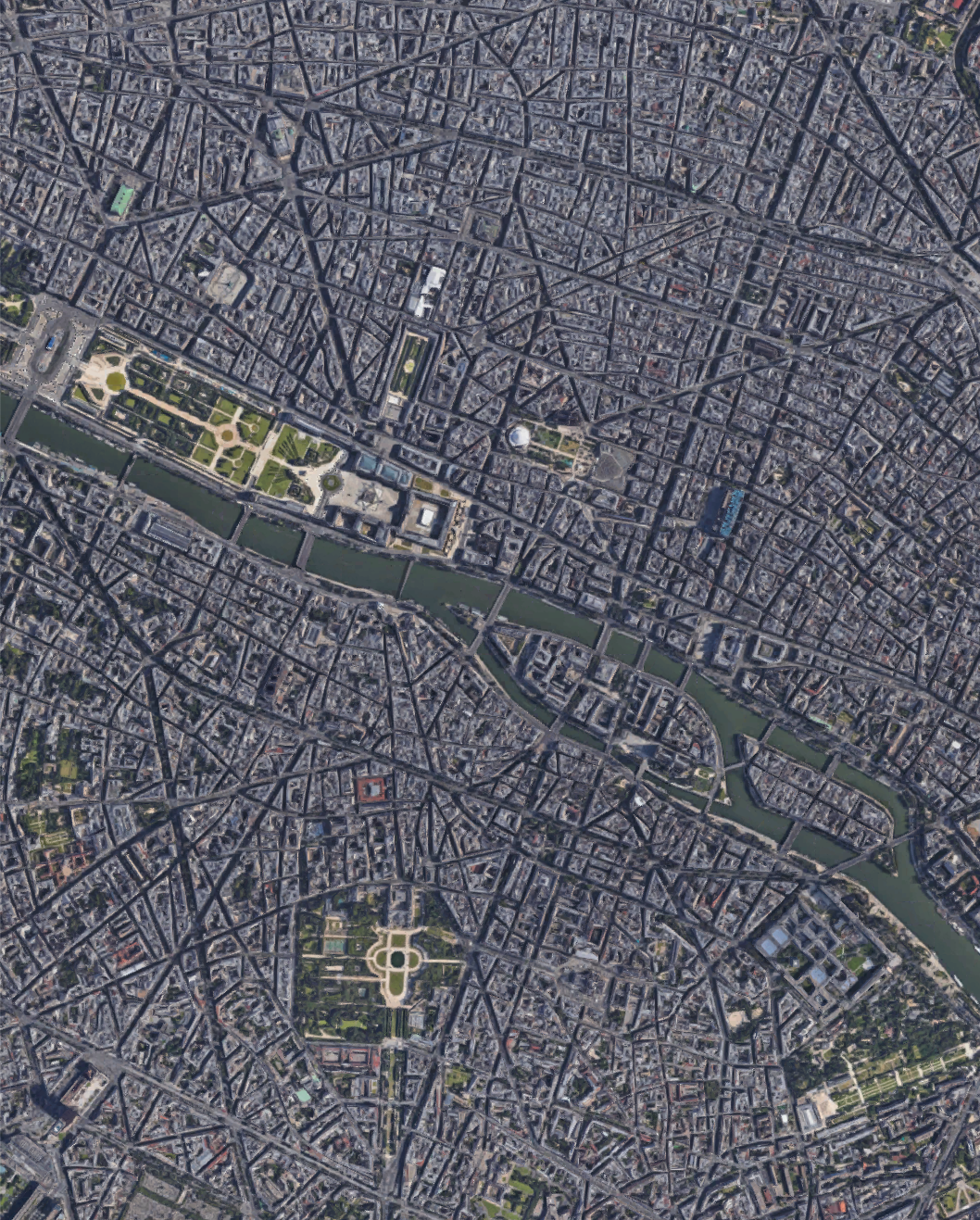}
\end{minipage}
  \caption{Views of the same geographical location (Paris, France). \textbf{Left:} OpenStreetMap.\\\textbf{Right:} Google Satellite Image.}
    \label{fig:paris:map}
\end{figure}

Coming back to the technical aspect, the main challenge of the project is to disentangle geographical information from other non-content domain variables.  We observe that the main non-content-related variations among maps arise from their artistic characteristics such as colors, symbols and legend styles. On the other hand, the underlying geographic information they are encoding remains the same. For example, although the map from the OpenStreetMap (Fig.~\ref{fig:paris:map}) and the satellite image from Sentinel (Fig.~\ref{fig:paris:map}) manifest different visual representation, they encode the same underlying geographic information such as geometric relations (road intersections, building arrangements) as they depict the same location.  

Unlike most of the current approaches (See Sec.~\ref{sec:relatedwork}), we focus on the encoding of the less-intuitive information about the underlying geographic relations with disentangled hidden factors.  In our project, we question what is the geographic information ("content") as opposed to the way they are visually represented ("style") and explore the VAE-variant model with adversarial training to disentangle those two main components of a map.

\subsection{Hypothesis}
\subsubsection{$\beta$- VAE with Adversarial Training}
\begin{figure}[ht]
\centering
\begin{minipage}{.35\linewidth}
  \includegraphics[width=\linewidth]{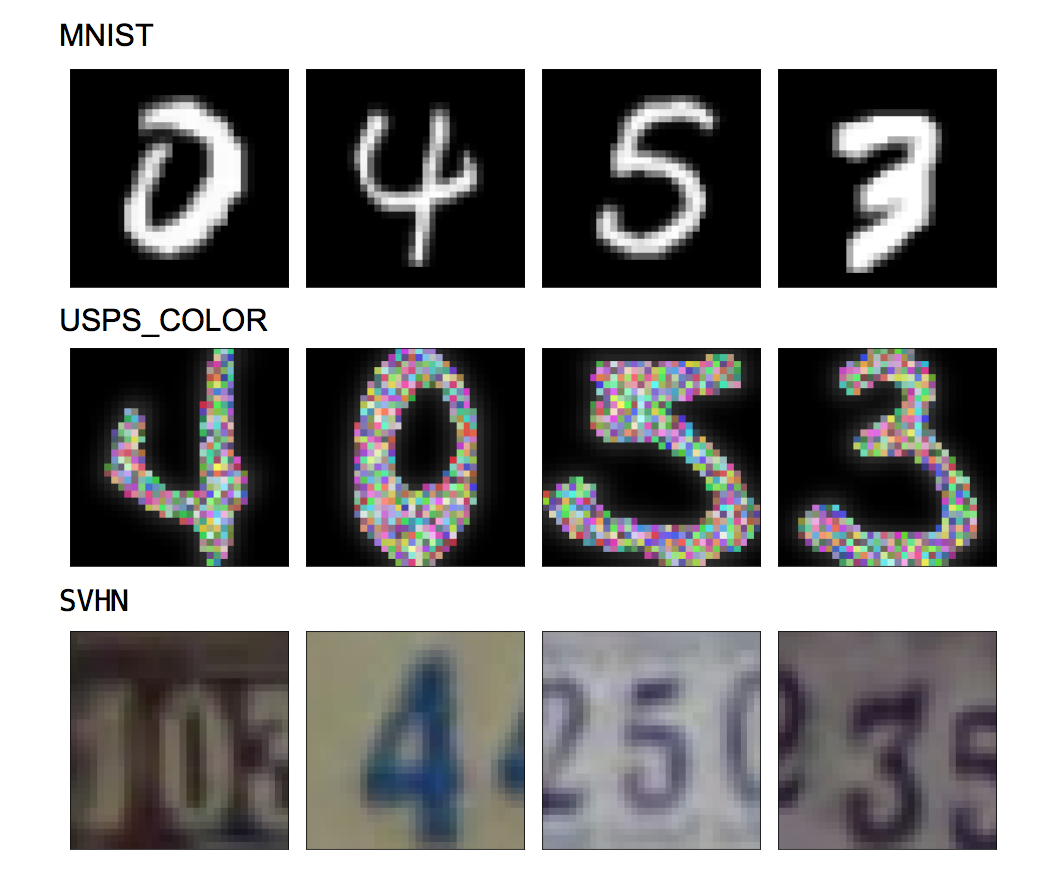}
  \caption{Digits datasets.}
  \label{digits}
\end{minipage}
\hspace{.05\linewidth}
\begin{minipage}{.45\linewidth}
  \includegraphics[width=\linewidth]{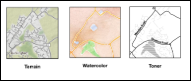}
  \caption{The same geographic area shown in 3 different styles : Terrain, Watercolor and Toner.}
  \label{maps}
\end{minipage}
\end{figure}

Driven by the question of extracting geographic information that is invariant to maps' artistic, visual representation, we set up proof-of-concept experiments on two collections of image triplets (See Fig. \ref{digits} and Fig. \ref{maps}). Both collections contain image triples that contain the same "content", yet appear visually different (i.e., have different "styles"). The first collection is generated from MNIST \cite{lecun-mnisthandwrittendigit-2010}, Street View House Number dataset \cite{svhn-data}, and modified USPS Digit dataset \cite{goodfellow2013multidigit}.

The second collection is produced from the Stamen Map\footnote{http://maps.stamen.com}, and its triplets are map tiles at the same geographic location and resolution (i.e., same latitude, longitude and zoom level), but with different artistic styles.  Fig. \ref{maps} shows an example of such triplet, with styles \texttt{Terrain}, \texttt{Watercolor}, and \texttt{Toner}.  More details on these datasets will be discussed in Sec.~\ref{sec:datasets}. 

We propose the second collection as a  platform to test whether a model has successfully learned to disentangle the essence of geographic information from other varying, visual aspects of a map.  We hypothesize that a combination of $\beta$-VAE-like architecture \cite{HigginsMPBGBML17} and adversarial training will achieve both (1) the disentanglement of the geographic information and artistic styles and (2) generation of new map tiles by composing the encoded geographic information with any artistic style.  We conjecture that augmenting a style classifier as a discriminator (i.e., adversary) of the encoder will help the separation of "content" and "style" information in the latent space. Fig.~\ref{fig:baseline} offers an overview of our model that reflects these hypotheses. A detailed description of the proposed architecture is discussed in Sec.~\ref{sec:model}.

\section{Background and Related Work} \label{sec:relatedwork}

Previous works on image registration in computer vision and GIS have explored image warping techniques such as SIFT~\cite{lowe1999} which is based on estimating "best" transformations from matching points and features, as well as physics-based models such as Demon's algorithm~\cite{pennec:demon} which morphs an image (often called the \textit{source} to its reference image, \textit{target}).  More recently, Deep Neural Network architectures have advanced the state of the art by, for instance, estimating matching features or modeling more complicated mapping from the source and the target images. 

A recent work from Liu \textit{et al.}~\cite{NIPS2018_7525}, proposes a unified feature disentanglement network (UFDN) that performs feature disentanglement and translation across multiple data domains. The model consists of encoder that takes as inputs data from distinct domains (ex. photo, sketch and paint images are considered as 3 different domains). The encoder learns a domain-invariant latent feature representation through adversarial training. It tries to confuse an adversary from correctly predicting a true domain label and as a result domain information is erased from the \textit{z} representation. The input's latent representation is concatenated to its domain vector (starts as random vectors).  The combine representation is passed to a generative adversarial network that aims to reconstruct the original image.

The model's performance is satisfactory, however, its architecture involves training a generative adversarial network that is both computationally expensive and technically hard. Hence, a simpler model that could perform effective multi-domain feature disentanglement and domain translation without the use of GAN would be desirable. 
\label{gen_inst}

\section{Datasets} \label{sec:datasets}

\subsection{Digits}
The MNIST, USPS, and Street View House Number (SVHN) datasets serve as our first collection of images with the same content (digits) but different style domains. The MNIST dataset contains 60000/10000 training/testing images and depicts white handwritten digits against a black background. The USPS dataset consists of 7291/2007 training/testing images and also pictures white digits against a dark background. To consider the style of the USPS dataset as a distinct style domain (and not the same as MNIST), we modify the images by adding Gaussian noise to pixels associated with their content information (digits). The SVHN contains 73257/26032 training/testing color house-number images with complex backgrounds and various illuminations. Even though each image in SVHN is associated with a single digit label, the images may contain multiple digits. To conduct our experiments we convert all images to  match 3x32x32 dimensions. 

\subsection{Maps: Stamen Dataset}
We created a collection of map tiles with three different base map styles (`Terrain’, `Toner’, and `Water Color’) using Stamen Map's API. We queried each map tile using the standard ``xyz" convention, and stored them with the naming convention of ``{x}-{y}-{z}.png".  Each map tile is an RGB image of size 256x256 pixels. Our dataset consists of 3 x 1411 tiles of Las Vegas (USA),3 x 2338 images of Shanghai (China), and 3 x 1270 map tiles of Paris (France). In addition to the map-style image tiles, we additionally collected satellite images from ESRI (`'EsriImagery"). We have collected 1200 (256x256) satellite image tiles for each city with the consistent zoom-level as for the Stamen map tiles. For our experiments, we used images from Paris and Las Vegas.  We believe this dataset adds a new flavor to the collection of datasets that the machine learning community often uses for benchmarking different models. In addition to geometric shapes (eg. digit datasets like MNIST, USPS, etc), faces (eg. CelabA), common objects (eg. COCO, ImageNet, Fashion MNIST), it introduces datasets regarding geospatial information.

\begin{figure}[ht]
  \centering
  \includegraphics[scale = 0.5]{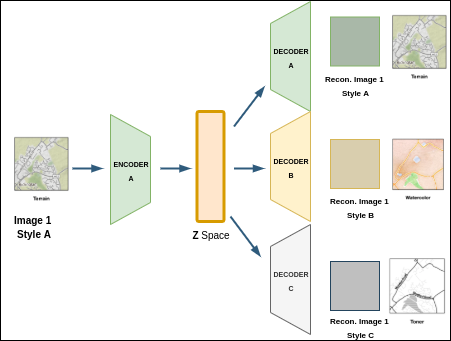}
  \caption{The image with the style label A passes only through the Encoder with the label A. The representation produced by the Encoder A is fed to all of the decoders in the model (Decoders A,B and C). The reconstruction loss is calculated by comparing the output of each decoder to the ground truth images for each of the style (A,B and C). The whole model consists of 3 encoders and 3 decoders.}
    \label{fig:baseline}
\end{figure}
\section{Architectures and Cost Functions } \label{sec:model}
In this study we experiment with two $\beta$-VAE~\cite{Kingma:2014} based models to achieve disentangled representation of geo-specific information.

\subsection{Baseline: Domain Specific Encoders-Decoders Networks}

As our baseline, we train a Domain-Specific Encoders-Decoders Network (DSED). The input to the model consists of triples of images. The images in the triple have the same content (digit or map of the same location) but come with different styles. Each of the pictures passes through a specific encoder: image with style label A passes through Encoder A, image with style label B passes through encoder B, etc. The encoders produce latent representations of their inputs, each of the learned representations serves as an input to all of the decoders in the model (See Fig.~\ref{fig:baseline} for details).

The DSED encourages the encoders to learn style invariant representations of inputs. Simultaneously, the model enforces the decoders in the network to store style information in their weights. The outputs produced by each of the encoder-decoder combinations are compared to the ground truth images in the input triplet. To train the model, we update the encoders with gradients from \textit{Mean Square Error (MSE)} (reconstruction loss) and \textit{Kullback-Leibler divergence} (penalizes deviation of latent feature from the prior distribution $p(z)$. The decoders are updated with gradients from reconstruction errors. The choice of \textit{MSE} as the reconstruction loss implies, our decoder is modeling the parameters for a (multivariate) Gaussian distribution over the image domain with the assumption that each pixel are independent from each other. 

 \begin{equation}
   L_{DSED}(\phi, \beta) =  MSE  +  \beta  D_{KL} (q_{\phi}(z_{i}|x_{i}) || (p_{\theta}(z_{i}|x_{i}) )
   \end{equation}
\begin{figure}[ht]
  \centering
  \includegraphics[width=10cm]{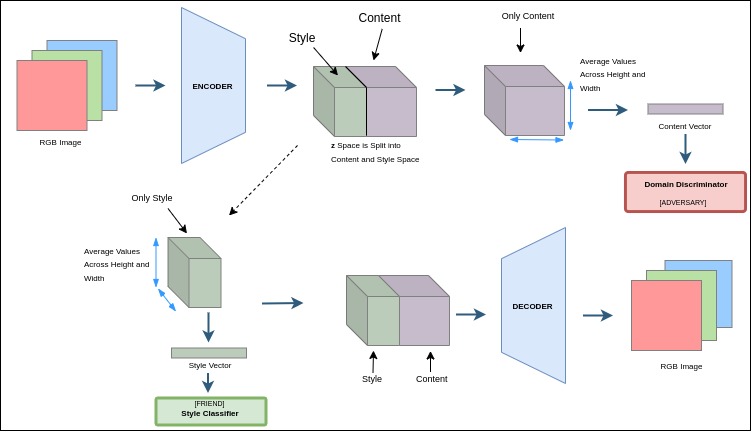}
  \caption{Overview of our $\beta$-VAE with a Friend and an Enemy Networks.}
     \label{fig:fen}
\end{figure}

\subsection{  $\beta$-VAE with Friend and Enemy Networks (FENs)}

$\beta$-VAE with FENs learns the disentangle representation of the latent space in a semi-supervised manner. As depicted in Fig.~\ref{fig:fen}, the network consists of a variational autoencoder, a domain discriminator (Enemy) and a domain classifier (Friend). The encoder E takes the image  $x_{j}$ as input and obtains its latent space representation. To be more specific, the encoder consists of four convolutional layers, the last of the layers outputs 64 x 4 x 4 rectangular parallel-piped. A  32-dimensional latent vector is obtained by averaging values of each filter before applying the reparametrization trick. Due to the use of $\beta$-VAE, we expect that the model learns a disentangled representation of the data. However, $\beta$-VAE does not guarantee the spacial disentanglement between the content and the style features in the latent space.

To achieve the above-mentioned objective, we add two single-layer neural networks to the model: Friend and Enemy. We select some of the latent dimensions to encode the content information ("content vector") while the rest encodes the style features ("style vector"). The relative size of the vectors depends on the relative complexity of the "style" and "content". In our experiments, we choose a smaller size vector for content when working with digits dataset (12:20) and a bigger size for content in the case of the maps dataset (20:12). The "content vector" is passed to the Enemy network that tries to classify the style of an image. The encoder E's objective is to confuse the Enemy network from predicting the correct style domain. Since we wish that the "content vector" does not contain any information about the style, we aim to maximize the cross-entropy loss of the Enemy network. 

To encourage encoding of the style information in our chosen dimensions we rely on the Friend network. The Friend network objective is to correctly classify the style of an image based on its corresponding "style vector". The encoder E aims to help the Friend network to produce correct labels for styles and minimize Friend's cross-entropy loss. Due to the antagonistic nature of the Friend and Enemy network consider only of these network losses per model's iteration.

The cost function during even iterations:
 \begin{equation}
 \begin{aligned}
   L_{FENs_{even}}(\phi, \beta, \gamma) = MSE  +  \beta \sum_{i=1}^{32} D_{KL} (q_{\phi}(z_{i}|x_{i}) || (p_{\theta}(z_{i}|x_{i}) )\\
           - \gamma \sum_{i=1}^{3} (y*log(y_{i}')+ (1-y)*log(1-y_{i}'))  
\end{aligned}
 \end{equation}
 The cost function during odd iterations:
 \begin{equation}
  \begin{aligned}
   L_{FENs_{odd}}(\phi, \beta, \gamma) =  MSE  +  \beta \sum_{i=1}^{32} D_{KL} (q_{\phi}(z_{i}|x_{i}) || (p_{\theta}(z_{i}|x_{i}) )\\
          +  \gamma \sum_{i=1}^{3} (y*log(y_{i}')+ (1-y)*log(1-y_{i}'))  
\end{aligned}
 \end{equation}
\section{Experiments }
In this section, we first present the results obtained during the testing of our baseline model on the digits collection. Next, we describe the results we achieved while training different variants of the proposed $\beta$-VAE FENs model on both digits and maps datasets and discuss the challenges in finding appropriate values for their hyper-parameters.

\begin{figure}[!htb]
\centering
\begin{minipage}{.32\linewidth}
  \includegraphics[width=3.8cm]{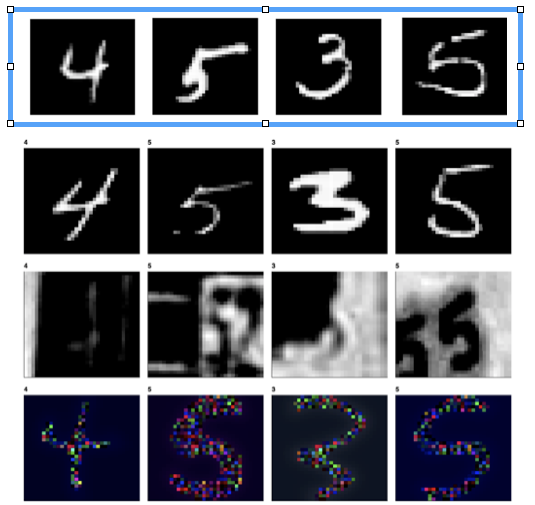}

\end{minipage}
\begin{minipage}{.32\linewidth}
  \includegraphics[width=3.8cm]{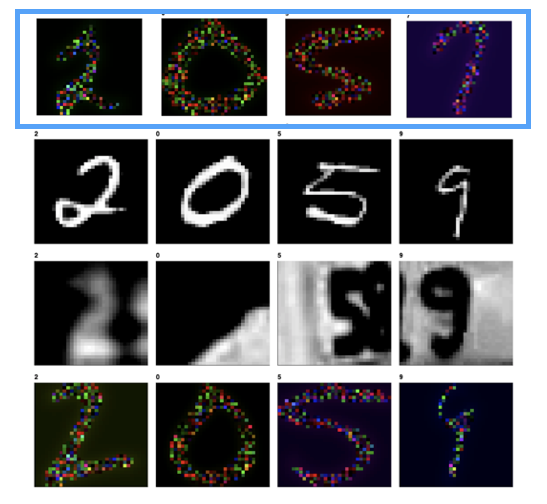}

\end{minipage}
\begin{minipage}{.32\linewidth}
  \includegraphics[width=3.8cm]{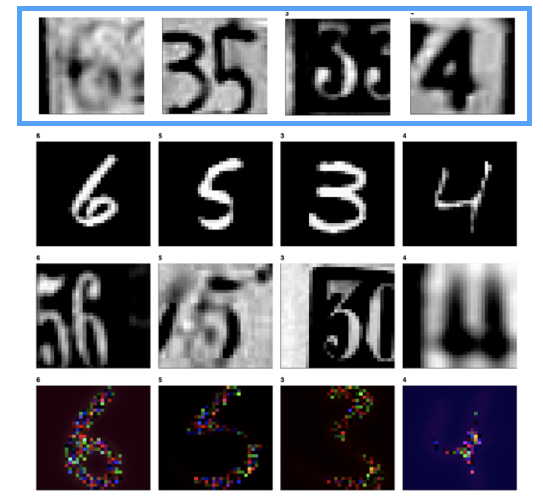}

\end{minipage}
\caption{ Examples of results. The images in the top row (boxed) correspond to the inputs to the network. \textbf{Left:} MNIST images in the top row are encoded and passed through the network decoders. The second row shows results of decoding the images with MNIST style decoder, the third row shows the results of decoding them with SVHN decoder and the fourth row consists of results of passing the latent representation of the input images through the USPS with Gaussian Noise style decoder. \\
\textbf{Middle: } As described above but the input images are from USPS with Gaussian Noise dataset. \\
\textbf{Right:} As described above but the input images are from SVHN dataset.}
\label{fig:exp:baseline}
\end{figure}

\subsection{Baseline: Domain Specific Encoders-Decoders Networks}

Our baseline model successfully learns the latent factors for the digits collection. For each encoder-decoder pair in a triplet, we used the same VAE architecture as proposed in the original $\beta$-VAE model. Our best model used the \textit{z} space of (1,32) and $\beta$ of 5.0. We split the digit datasets into training and testing in a 3:1 ratio, and at each iteration batch size of (12x3) (3 for the number of styles) images were used to compute the gradients. We assumed that the decoders parameterize a Gaussian distribution over the output (RGB) space and thus used the MSE as our reconstruction loss. The total loss was computed by adding the standard total KL divergence (averaged over the batch size) and the reconstruction loss.  Our results in Fig.~\ref{fig:exp:baseline} show the outputs from each decoder when we provide a specific style of input.  The images verify that the decoder correctly learns the distinct style domains from the digit collection and successfully generates new images of digits with desired styles.

\subsection{  $\beta$-VAE with Friend and Enemy Networks (FENs)}
\subsubsection{Training Results}
\begin{figure}[ht]
\centering
\begin{minipage}{.45\linewidth}
  \includegraphics[width=4.5cm]{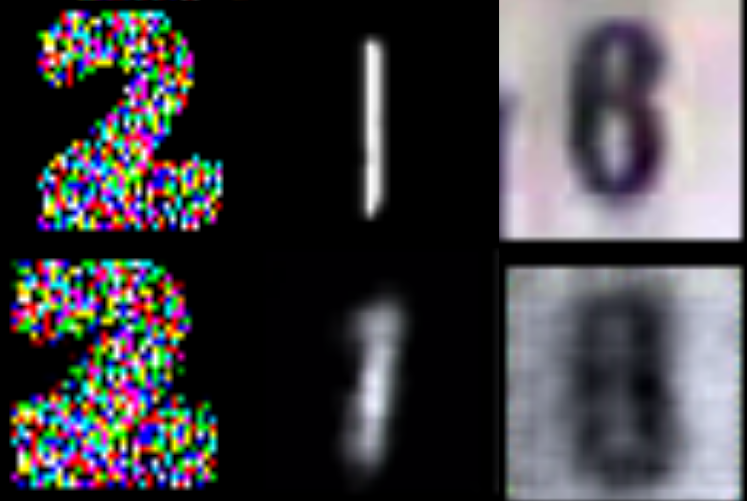}

\end{minipage}
\begin{minipage}{.45\linewidth}
  \includegraphics[width=4.5cm]{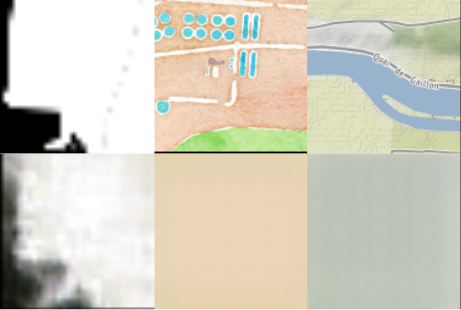}
\end{minipage}
\caption{Examples of results. The ground truth images are in the first row, the reconstructions are in the second row. \textbf{Left:} The reconstruction results achieved during training of $\beta$-VAE with FENs on MNIST,UPSP with Gaussian Noise and SVHN datasets. \textbf{Right:} The reconstruction results for the MAPs dataset. From left: Toner, Watercolor, and Terrain styles.}
\label{fig:fen_recon}
\end{figure}

We run experiments with the proposed $\beta$-VAE FENs architecture on the two image collections (Sec.~\ref{sec:datasets}) with various hyper-parameter settings and \textit{z}-space sizes (32,64,128,512) and dimensions (2D,3D ex.32x4x4). First, on the digit dataset, we investigated the performance of the model with one-dimensional latent space of size 1x32, and achieved the reconstruction results shown in Fig.~\ref{fig:fen_recon}, left side. The figure shows images produced by a model trained with Adam optimizer, $\beta = 4.0$, $ \gamma = 100$, and  learning rate of 0.001 for 650 epochs.  On the right side of Fig.~\ref{fig:fen_recon}, we present the reconstruction achieved by $\beta$-VAE FENs with one dimensional latent space of size 1x128 on the maps dataset. This model was trained with Adam optimizer, $\beta = 4.0$, $ \gamma = 2.0$, and  learning rate of 0.001 for 950 epochs. It learned to reconstruct the images from the 'Toner' style domain. For 'Watercolor' and 'Terrain' domain the model was only able to capture their monochromatic color representation. The values of loss functions are provided in Table 1.

\begin{table}
     \caption{Table 1. The loss functions for our best performing $\beta$-VAE with FENs model for each of the datasets.}
     % \label{sample-table}
    \begin{center}
     \begin{tabular}{ c c c c c c c }
     % \begin{tabular}{ c | c | c | c | c| c | c }
     \toprule
    $\beta$-VAE FENs& Reconst. Loss (MSE) & KLD Loss & Friend Loss & Enemy Loss &$\beta$&$\gamma$\\  
    \midrule
    Digits Dataset &2028.71& 417.30&172.76&267.95&4.0&100.00\\
     \midrule
    Maps Dataset &109.39 &20.06&173.79 & 267.40&4.0&2.0\\
     \bottomrule

    \end{tabular}
    
    \end{center}
\end{table}

We also experimented with various settings of hyperparameters to train the FEN model and learned the challenges of finding the right balance between the weights in our loss function (ie. $\beta$, $\gamma$) as well as the learning rate.  As discussed in [4], the choice of `'best" $\beta$ is dependent on the complexity of each dataset, and thus needs to be experimentally determined. In addition to the balance between the KL divergence and the reconstruction loss, our model requires an extra juggling with the loss from friend/enemy classifier, controlled by $\gamma$.

\begin{figure}[ht]
  \centering
  \includegraphics[width=14cm]{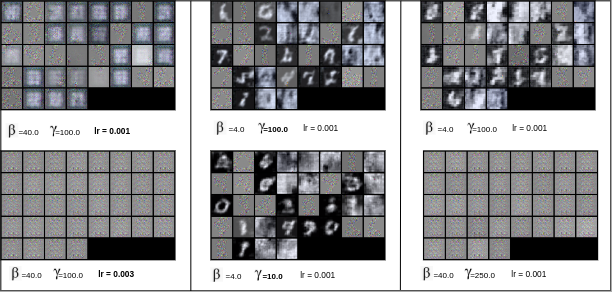}
  \caption{Results of training digits dataset with the same $\beta$-VAE with FENs architecture and batch size (36) for 450 epochs. The results vary due to different hyper-parameters settings.}
     \label{fig:fen_params}
\end{figure}

Fig.~\ref{fig:fen_params} demonstrates the reconstruction results after the FEN models are trained for 450 epochs with the specified $\beta$, $\gamma$ and learning rate. In the first column, we observe the effect of the learning rate when the other hyperparameters are the same. Comparing the first two images in the first row reveals that a proper ratio of $\beta$ and $gamma$ is necessary to train a good model.  On the other hand, the comparison of the left-top figure with the figure in the second row, second column tells us that it is not just the ratio of the $beta$ and $gamma$, but the actual magnitude of the parameters. We hypothesize that this is due to how they affect the total loss while interacting with the reconstruction loss (which are often several magnitudes larger at the beginning of the training).

\subsubsection{Testing Regime}
In order to test $\beta$-VAE with FEN model, we pass through the network 2 images. We keep the "content vector" from the first image (image 1) and concatenate it to "style vector " representing the style of second image (image 2). Then we passed the newly created representation to the decoder to create a style-transferred image. We could not generate good style transfer for our model due to insufficient disentanglement in latent space. Extra tuning of model parameters will help us obtain better result but due to the limitation of time and computational resources we will leave it for future work.  

\section{Future Work}

First, we plan to finish the style transfer of digits with FENs. We need more computational resources to tune parameters $\beta$ ,$\gamma$ in objective functions. We also need to play with different training scheme apply enemy loss and friend loss to maximize the effect.

\bibliographystyle{plain}
\bibliography{mainbib}

\end{document}